\documentclass{article}

\PassOptionsToPackage{numbers, compress}{natbib}
\usepackage[dblblindworkshop, final]{neurips_2026}

\usepackage[utf8]{inputenc} \usepackage[T1]{fontenc}    \usepackage{hyperref}       \usepackage{url}            \usepackage{booktabs}       \usepackage{amsfonts}       \usepackage{nicefrac}       \usepackage{microtype}      \usepackage{xcolor}         \usepackage{kotex}
\usepackage{booktabs}
\usepackage{amsmath}
\usepackage{graphicx}
\usepackage{float}

\definecolor{Rhodamine}{RGB}{227, 0, 122}
\definecolor{BrightGreen}{HTML}{00AB4E}
\definecolor{cagpink}{RGB}{255,51,153}
\definecolor{cagpurple}{RGB}{128,0,128}
\definecolor{blue}{RGB}{0, 102, 204}

\title{Completion Aware Guidance for World Action Models}
\workshoptitle{Robot Learning with World Models: Capabilities, Frontiers, and Challenges}
\author{
  Seungyeon Kim$^{1}$ \quad
  Junhoo Lee$^{2}$ \quad
  Baekseung Kim$^{1}$ \quad
  Minkyu Kim$^{1}$ \quad
  Nojun Kwak$^{1}$ \\[4pt]
  \normalfont $^{1}$Seoul National University \quad
  \normalfont $^{2}$KAIST \\[4pt]
  \texttt{\{syeonkim,baekseung.kim,minkyu.kim,nojunk\}@snu.ac.kr} \\
  \texttt{junhoo.lee@kaist.ac.kr}
}

\begin{document}
\maketitle

\begin{abstract}
World Action Models (WAMs) predict visual futures and robot actions, yet they remain susceptible to \emph{task-incomplete imagination}, where plausible, action-consistent predictions omit the transition needed for task completion. In this paper, we show that this failure is not inherent to the world model backbone, but emerges when adapted for short-chunk control, which can repeatedly favor plausible local continuations over task-completing transitions. To address this, we introduce Completion Aware Guidance (CAG), a training-free sampling method that guides generation toward task completion. Across representative WAMs, CAG improves success from 64\% to 70\% on a RoboTwin 2.0 subset and from 69\% to 75\% in zero-shot simulation, while reducing task-incomplete imagination from 79\% to 40\%.
\end{abstract}

\section{Introduction}
Robot tasks unfold through changes in the world state, and therefore require temporal understanding of what is likely to happen after an action is executed. Vision--language--action policies \citep{brohan2022rt,brohan2023rt,kim2024openvla,black2024pi_0,intelligence2025pi_} predict actions from the current observation, yet the resulting future changes are difficult to infer from {{\color{black}observed}} images alone. World Action Models (WAMs) address this limitation by using video generative models to jointly predict a visual future and the actions that realize it \citep{ye2026world,yuan2026fast,shen2026world}. However, generating visual futures with aligned actions does not guarantee task success.

\begin{figure}[t]
    \centering
    \includegraphics[width=\linewidth]{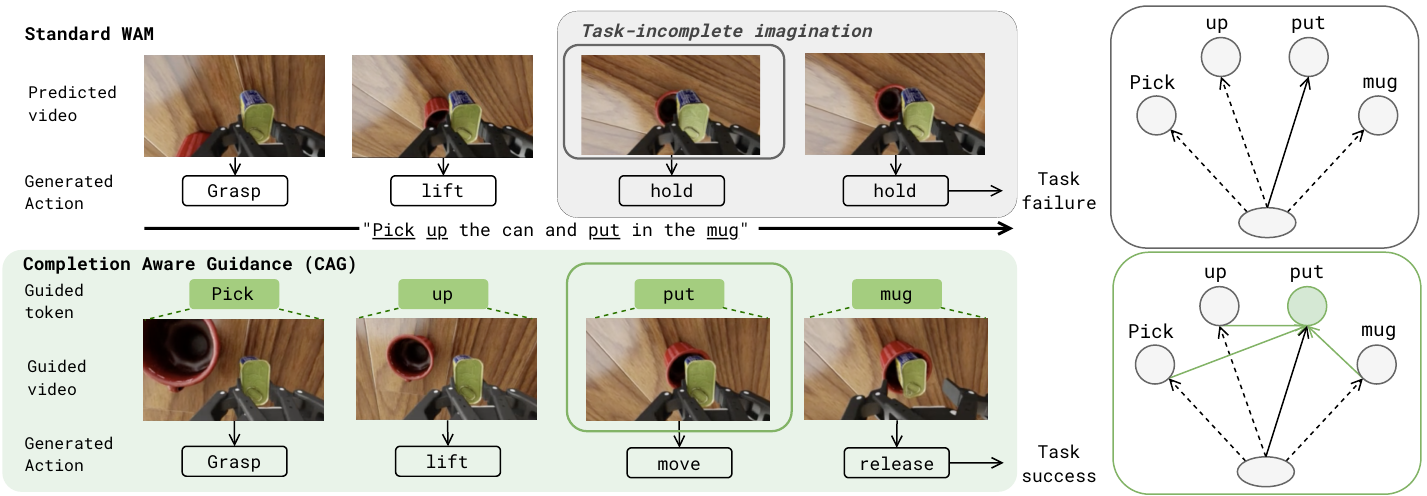}
    \vspace{-4mm}
    \caption{\textbf{Completion Aware Guidance.} A standard WAM produces a visually plausible, action-consistent continuation while omitting the transition required for task completion. CAG strengthens instruction conditioning relevant to the current prediction and recovers the transition during control.}
    \label{fig:overview}
    \vspace{-3mm}
\end{figure}

Interestingly, as shown in Fig.~\ref{fig:overview}, we find that WAMs exhibit \emph{task-incomplete imagination}, generating visually coherent, instruction-relevant futures while omitting the state transitions required for task completion. {{\color{black}For example, given a {{\color{black}pick-and-place}} instruction,}} the imagined video may move an object above its target but continue holding it instead of releasing it, and the predicted actions follow this future, resulting in task failure. This differs from a {{\color{black}corrupted video}} or {{\color{black}video--action misalignment}} {{\color{black}in that}} both modalities agree on a {{\color{black}plausible continuation}} that does not complete the task.

In this paper, we ask why WAMs exhibit task-incomplete imagination and how this failure can be addressed. We argue that it arises from an objective mismatch between the fixed-horizon video-generation objective {{\color{black}for video-backbone pretraining}} and the short-horizon video-action prediction objective {{\color{black}for WAM fine-tuning}}. {{\color{black}Under the former objective,}} video diffusion models jointly predict an instruction-following trajectory over a fixed horizon, whereas WAMs predict {{\color{black}video-action chunks}} that are executed and regenerated from the {{\color{black}observation}}. {{\color{black}Because the WAM objective does not require the current phase to be completed within {{\color{black}each horizon}}, the state transition required for task completion can be repeatedly deferred.}}
As a result, although the video backbone can generate instruction-following trajectories, this capability is not reliably elicited {{\color{black}when needed}}.

To address this mismatch, we introduce \textbf{Completion Aware Guidance (CAG)}, a training-free sampling method for WAMs. CAG strengthens the conditioning of instruction tokens most relevant to the current video--action prediction, steering the current chunk toward the required state transition without retraining the WAM. Across DreamZero and Fast-WAM \citep{ye2026world,yuan2026fast}, CAG improves {{\color{black}the}} average task success {{\color{black}rate}} from {{\color{black}64.4\% to 70.0\%}} on a RoboTwin 2.0 subset and from 69\% to 75\% in zero-shot simulation. More importantly, in a representative manipulation setting exhibiting task-incomplete imagination, CAG further reduces the incidence of this failure from 79\% to 40\%.

\section{Related Work}
\vspace{-2mm}

\textbf{Vision-Language-Action Models.}
Generalist robot policies have largely been developed as vision--language--action models, which map visual observations and language instructions directly to actions or action chunks  \citep{brohan2022rt,brohan2023rt,kim2024openvla,black2024pi_0,intelligence2025pi_}. For temporally extended tasks, however, a global instruction may not specify the next semantic subtask or desired near-future state. Recent approaches~\citep{intelligence2026pi} therefore provide language subtask instructions or visual subgoal images to make task progress explicit \citep{ahn2022can,du2024video}. Yet specifying an intermediate target and reaching it are distinct problems~\citep{kim2026recursive}: external guidance does not ensure that the policy produces the state transition required to complete the subtask. 

\textbf{World Models and World Action Models.}
{{\color{black}Video generative models capture temporal dynamics in {{\color{black}observations}}, making them natural backbones for predictive world models in robot planning and control}} \citep{wu2023daydreamer,du2023learning,wu2024unleashing,cheang2024gr,hu2024video,zhou2024robodreamer,liang2024dreamitate,zhu2025irasim,agarwal2025cosmos}. WAMs extend this idea by coupling the imagined visual trajectory with {{\color{black}actions}}, allowing a generated future to act as an implicit temporal plan \citep{ye2026world,yuan2026fast,shen2026videovla,li2025unifiedvideoactionmodel,shen2026world}. {{\color{black}Prior works\citep{ye2026world,yuan2026fast,ye2026gigaworld,cai2026aha} do not isolate the case in which the imagined video and predicted actions remain mutually consistent and locally plausible while omitting a phase-completing transition. We characterize this failure as \emph{task-incomplete imagination} and mitigate it with training-free sampling guidance.}}

\section{Method}
\vspace{-2mm}
\textbf{Task-Incomplete Imagination.}
{{\color{black}As illustrated in Fig.~\ref{fig:overview}, WAMs can repeatedly produce locally reasonable video-action continuations while failing to progress to the transition required for the next task phase. We refer to this failure as \emph{task-incomplete imagination}; for example, the model may reach the target region yet continue holding the object instead of releasing it. Does this failure simply reflect a limitation of the underlying video diffusion backbone?}}

{{\color{black}Using the same video diffusion backbone, we compare its native generation horizon with the short chunk horizon used by the WAM. As shown in Fig.~\ref{fig:result_summary} (a), the native horizon produces the required completion event, whereas short-chunk autoregressive generation remains visually coherent but repeatedly omits it. This suggests that task-incomplete imagination is not an intrinsic limitation of the backbone, but arises from its adaptation to short-horizon control generation as follows:}}
\newsavebox{\cagDreamZeroTable}
\newlength{\cagResultHeight}
\newlength{\cagTitleHeight}

\sbox{\cagDreamZeroTable}{\renewcommand{\arraystretch}{1.4}\resizebox{0.31\textwidth}{!}{\begin{tabular}{lcc}
    \toprule
    Task & DreamZero~\cite{ye2026world} & CAG \\
    \midrule
    Cube to Bowl  & 90 & 92 \\
    Can to Mug    & 60 & 72 \\
    Banana to Bin & 56 & 60 \\
    \bottomrule
    \end{tabular}}}

\setlength{\cagResultHeight}{0.17\textwidth}
\setlength{\cagTitleHeight}{1.4\baselineskip}
\begin{figure*}[t]
\centering
\noindent

\begin{minipage}[t]{0.425\textwidth}
\vspace{0pt}
\centering

\parbox[t][\cagTitleHeight][t]{\linewidth}{\centering
  \mbox{\footnotesize\bfseries
  (a) Video completion across horizons ($\uparrow$)}}\par\nointerlineskip

\begin{minipage}[t][\cagResultHeight][c]{\linewidth}
\centering
\includegraphics[
  width=\linewidth,
  height=\cagResultHeight,
  keepaspectratio,
  trim=1pt 1pt 1pt 1pt,
  clip
]{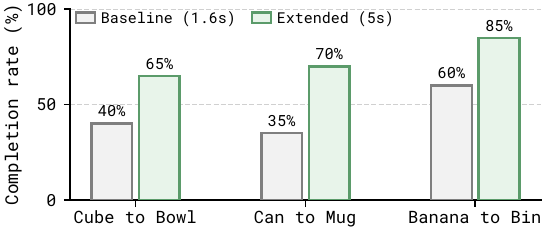}
\label{fig:result_summarya}
\end{minipage}

\end{minipage}\hfill\begin{minipage}[t]{0.31\textwidth}
\vspace{0pt}
\centering

\parbox[t][\cagTitleHeight][t]{\linewidth}{\centering
  \mbox{\footnotesize\bfseries (b) DreamZero zero-shot task success}}\par\nointerlineskip

\begin{minipage}[t][\cagResultHeight][c]{\linewidth}
\centering
\usebox{\cagDreamZeroTable}
\end{minipage}

\end{minipage}\hfill\begin{minipage}[t]{0.235\textwidth}
\vspace{0pt}
\centering

\parbox[t][\cagTitleHeight][t]{\linewidth}{\centering
  \mbox{\footnotesize\bfseries (c) Failure incidence ($\downarrow$)}
}\par\nointerlineskip

\begin{minipage}[t][\cagResultHeight][c]{\linewidth}
\centering
\includegraphics[
  width=\linewidth,
  height=\cagResultHeight,
  keepaspectratio,
  trim=1pt 1pt 1pt 1pt,
  clip
]{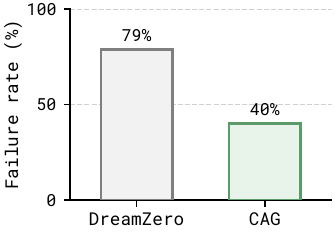}
\end{minipage}

\end{minipage}

\caption{
\textbf{Analysis of CAG on DreamZero.}
\textbf{(a)} Video completion rates under standard 1.6-second chunk generation and extended 5-second generation, showing that a longer generation horizon recovers task-completing transitions.
\textbf{(b)} Zero-shot task-success rates for DreamZero and CAG.
\textbf{(c)} Proportion of failed rollouts attributed to task-incomplete imagination,
which is reduced by CAG.}
\label{fig:result_summary}
\vspace{-3mm}
\end{figure*}

\begin{equation}
\underbrace{
(o_0,c)\longmapsto o_{1:T}
}_{\text{WM: task-level visual trajectory}}
\quad
\xrightarrow{\text{WAM adaptation}}
\quad
\underbrace{
(o_{\le t},q_t,c)\longmapsto
\bigl(o_{t+1:t+H},a_{t:t+H-1}\bigr)
}_{\text{WAM: local video--action continuation}}.
\label{eq:comparison}
\end{equation}

Here, $o_t$ denotes the observation at time $t$, $a_t$ the robot action,
$c$ the language instruction, $q_t$ the {{\color{black}proprioceptive state}},
$T$ the native video-generation horizon, and $H$ the WAM chunk horizon.
Once the task reaches a completion-relevant phase, the WAM must select the required
transition within a {{\color{black}short horizon}}. However, a holding or hovering continuation
can remain locally plausible and compatible with the instruction, providing no explicit
preference for realizing the completion in the current chunk. Repeated {{\color{black}generation}}
can therefore defer the required transition. This raises the central question: how can
we guide short-horizon generation to include the completion required by the current task phase?

\textbf{Completion Aware Guidance.}
Our algorithm, Completion Aware Guidance (CAG), tackles task-incomplete imagination as an elicitation problem. The video diffusion backbone represents the task-completing transition specified by the instruction, but short-horizon control can favor a {{\color{black}plausible continuation}} that preserves the pre-completion state. CAG biases this {{\color{black}selection}} toward a continuation that completes the current manipulation phase while remaining close to the {{\color{black}WAM prior}}.

Let $h_t=(o_{\leq t},q_t)$ denote the current WAM context and $z_t=(o_{t+1:t+H},a_{t:t+H-1})$ the predicted video-action chunk. We introduce a phase-completion indicator $E_t(z_t)\in\{0,1\}$, with $E_t(z_t)=1$ when the chunk contains the transition that completes the active task phase. Let $p_\theta(\cdot\mid h_t,c)$ denote the chunk distribution of the WAM with parameters $\theta$. The ideal completion aware distribution is

\begin{equation}
    p_t^\star
    =
    \arg\max_{p(\cdot\mid h_t,c)}
    \;
    \mathbb E_{z_t\sim p(\cdot\mid h_t,c)}
    [E_t(z_t)]
    -
    \lambda D_{\mathrm{KL}}\!\left(
    p(\cdot\mid h_t,c)
    \,\Vert\,
    p_\theta(\cdot\mid h_t,c)
    \right).
    \label{eq:ideal_completion_distribution}
\end{equation}

The first term favors video-action chunks that complete the active phase, while the KL regularizer keeps the guided distribution close to the original WAM prior. Here, $p(\cdot\mid h_t,c)$ denotes the candidate chunk distribution and $\lambda>0$ controls the strength of this regularization. The frozen model does not allow direct control over \(p_t^\star\). Its instruction signal enters video-action generation through cross-attention. At denoising layer \(\ell\), video-action queries \(Q^{va}_{t,\ell}\) receive the conditional context

\begin{equation}
    C^{va\leftarrow c}_{t,\ell}
    =
    \operatorname{softmax}\!\left(
    \frac{Q^{va}_{t,\ell}(K^c_\ell)^\top}{\sqrt d}
    \right)V^c_\ell,
    \label{eq:cross_attention_path}
\end{equation}

where $K_\ell^c$ and $V_\ell^c$ are the keys and values of the instruction tokens,
and $d$ is the attention key dimension. This pathway identifies where the instruction components used by the current video-action generation enter the model. We restrict the intervention to this pathway. For a token-wise scaling vector $\gamma_t$, we modify the key and value associated with each instruction token $j$ as

\begin{equation}
    \widetilde K_{\ell,j}^c
    =
    \gamma_{t,j}K_{\ell,j}^c,
    \qquad
    \widetilde V_{\ell,j}^c
    =
    \gamma_{t,j}V_{\ell,j}^c,
    \label{eq:kv_scaling}
\end{equation}
which induces the conditional chunk distribution \(p_\theta^{\gamma_t}(z_t\mid h_t,c)\). Ideal intervention is
\begin{equation}
    \gamma_t^\star
    =
    \arg\max_{\gamma_t}
    J_t(\gamma_t),
    \qquad
    J_t(\gamma_t)
    =
    \mathbb E_{z_t\sim p_\theta^{\gamma_t}(\cdot\mid h_t,c)}
    [E_t(z_t)]
    -
    \lambda D_{\mathrm{KL}}\!\left(
    p_\theta^{\gamma_t}(\cdot\mid h_t,c)
    \,\Vert\,
    p_\theta(\cdot\mid h_t,c)
    \right).
    \label{eq:ideal_kv_intervention}
\end{equation}

Around the original conditioning \(\gamma_t=\mathbf1\), the effect of an intervention is locally characterized by
\begin{equation}
    J_t(\gamma_t)-J_t(\mathbf1)
    \approx
    \left\langle
    g_t,\,
    \gamma_t-\mathbf1
    \right\rangle,
    \qquad
    g_{t,j}
    =
    \left.
    \frac{\partial J_t}{\partial\gamma_{t,j}}
    \right|_{\gamma_t=\mathbf1}.
    \label{eq:ideal_sensitivity}
\end{equation}
The vector \(g_t\) specifies which instruction components should be strengthened to increase completion elicitation. However, neither the phase-completion indicator \(E_t\) nor the resulting sensitivity \(g_t\) is available during sampling. CAG instead uses the cross-attention pathway to identify the instruction components that the current video-action generation already relies on.  For each instruction token $w_j$, we aggregate its attention affinity over video-action queries and a set of selected denoising layers $\mathcal{L}$,
\begin{equation}
    r_{t,j}
    =
    \frac{1}{|\mathcal{L}|}
    \sum_{\ell \in \mathcal{L}}
    \frac{1}{|Q^{va}_{t,\ell}|}
    \sum_{\mathbf{u} \in Q^{va}_{t,\ell}}
    \left[
    \operatorname{softmax}
    \left(
    \frac{\mathbf{u}(K_\ell^c)^\top}{\sqrt{d}}
    \right)
    \right]_j .
\end{equation}

CAG converts this observable signal into a bounded conditional intervention,
\begin{equation}
    \widehat\gamma_{t,j}
    =
    1+\alpha\,\psi(r_{t,j}),
    \label{eq:cag_rule}
\end{equation}

where \(\psi\) is a sparse binary gating function. Applying Eq.~\eqref{eq:cag_rule} through Eq.~\eqref{eq:kv_scaling} strengthens the instruction signal already used for the current action, biasing the next video-action chunk toward the completion 
required by the active phase.

\section{Experiments}
\vspace{-2mm}
\textbf{Setting.}
CAG is a training-free inference-time method applicable to WAMs built on video diffusion backbones. To test its generality across different WAM designs, we evaluate it on DreamZero and Fast-WAM, which respectively use explicit future generation and implicit world modeling for action prediction. We report task success over 50 rollouts on each of the three official DreamZero DROID simulation scenes and 30 rollouts per task  on nine non-saturated RoboTwin 2.0 tasks with published success rates below 90\% under the Fast-WAM Random setting. We follow the corresponding evaluation protocols, changing only inference-time guidance between the baseline and CAG, and fix $\alpha=0.15$ across all tasks and backbones without per-task tuning.
\begin{table}[t]
\centering
\small
\setlength{\tabcolsep}{6pt}
\caption{\textbf{RoboTwin 2.0 results.}
CAG improves overall task success on non-saturated tasks.}
\label{tab:robotwin}
\begin{tabular*}{\textwidth}
{@{\extracolsep{\fill}}lcccc}
\toprule
Task & $\pi_{0.5}$~\cite{intelligence2025pi_} & GO-1~\cite{bu2025agibot} & Fast-WAM~\cite{yuan2026fast} & CAG \\
\midrule
Open Microwave
    & \textbf{37} & 14 & 33 & 30 \\
Hanging Mug
    & 3 & 0 & 60 & \textbf{63} \\
Turn Switch
    & 6 & 30 & 53 & \textbf{67} \\
Place Can Basket
    & 25 & 37 & \textbf{63} & 53 \\
Move Stapler Pad
    & 18 & 4 & 57 & \textbf{73} \\
Stack Bowls Three
    & 35 & 7 & 67 & \textbf{87} \\
Pick Diverse Bottles
    & 3 & 56 & 80 & \textbf{83} \\
Place Mouse Pad
    & 26 & 10 & 80 & \textbf{83} \\
Place Object Basket
    & 36 & 49 & 87 & \textbf{90} \\
\bottomrule
\end{tabular*}
\end{table}

\sbox{\cagDreamZeroTable}{\resizebox{0.31\textwidth}{!}{\begin{tabular}{lcc}
    \toprule
    Task & DreamZero & CAG \\
    \midrule
    Cube to Bowl  & 90 & 92 \\
    Can to Mug    & 60 & 72 \\
    Banana to Bin & 56 & 60 \\
    \bottomrule
    \end{tabular}}}

\textbf{Results.}
CAG improves task success on both backbones. On the nine-task RoboTwin 2.0 subset, CAG increases average success from 64.4$\%$ to 70.0$\%$ (Tab.~\ref{tab:robotwin}). On three DreamZero zero-shot tasks, CAG improves average success from 69$\%$ to 75$\%$, with gains on every task (Fig.~\ref{fig:result_summary} (b)). The consistent gains across DreamZero and Fast-WAM suggest that CAG is not tied to explicit future-video generation, but generalizes across different ways of incorporating video-based world modeling into action prediction.

\textbf{Failure Analysis.}
To examine whether CAG directly addresses task-incomplete imagination, we analyze failure cases in the DreamZero zero-shot evaluation. As shown in Fig.~\ref{fig:result_summary} (c), the fraction attributable to task-incomplete imagination decreases from 79$\%$ to 40$\%$ with CAG, indicating that CAG gains from mitigating task-incomplete imagination.

\section{Conclusion}
\vspace{-2mm}
We identify task-incomplete imagination, in which a WAM produces a visually plausible, action-consistent continuation while omitting the state transition required to complete the current task phase. We attribute this failure to the mismatch between fixed-horizon video generation and short-horizon WAM control. CAG addresses this mismatch by strengthening instruction conditioning relevant to the current prediction, eliciting task-completing transitions during sampling. Across representative WAMs, CAG improves task success and reduces task-incomplete imagination without retraining.

\textbf{Limitations.}
CAG is currently limited to sampling-time intervention, and extending
completion aware guidance to WAM training is a promising direction for future work.

\begin{ack}
This work was supported by the Korean Government through
grants from IITP (RS-2021-II211343 and RS-2025-25442338), NRF (RS-2026-25500579) and NIPA (RS-2026-25621677).
\end{ack}

\bibliographystyle{unsrtnat}
\bibliography{nips}

\end{document}